\documentclass[5p]{elsarticle}
\usepackage{booktabs,shortvrb}
\usepackage{amsmath}
\usepackage{amsfonts}%
\usepackage{amssymb}%
\usepackage{booktabs}
\usepackage{url}
\usepackage{algpseudocode}
\usepackage{algorithm}

\makeatletter
    \def\ps@pprintTitle{%
      \let\@oddhead\@empty
      \let\@evenhead\@empty
      \let\@oddfoot\@empty
      \let\@evenfoot\@oddfoot
    }
\makeatother
\begin{document}

\begin{frontmatter}
\title{New results of ant algorithms for the Linear Ordering Problem}
\author[1]{Camelia-M. Pintea}\ead{cmpintea@yahoo.com}
\author[2]{Camelia Chira}\ead{chira@cs.ubbcluj.ro}
\author[2]{D. Dumitrescu}\ead{ddumitr@cs.ubbcluj.ro}
\address[1]{George Cosbuc N.College, Avram Iancu 70-72, Cluj-Napoca, Romania}
\address[2]{Department of Computer Science, Babes-Bolyai University, Cluj-Napoca 400084, Romania}

\begin{abstract}
Ant-based algorithms are successful tools for solving complex problems. One of these problems is the Linear Ordering Problem (LOP). The paper shows new results on some LOP instances, using  Ant Colony System (ACS) and the Step-Back Sensitive Ant Model (SB-SAM).
\end{abstract}

\end{frontmatter}

\section{Introduction}
Linear Ordering Problem (LOP) is an NP-hard problem \cite{Chanas1996, Karp1972}. In graph theory, LOP is searching for an acyclic tournament having the maximum sum of arc weights in a complete weighted graph. In other words, LOP is seeking a permutation of rows and columns in a given matrix of weights in order to maximize the sum of weights in the upper triangle \cite{Karp1972,Laguna1999}.

Several methods were proposed to solve the Linear Ordering Problem. Exact methods \cite{Mitchell2000,Reinelt1985,Reinelt1981} and heuristic methods were proposed to address LOP algorithms. Heuristics are more efficiently in finding near-optimal solutions to LOP: tabu search \cite{Garcia2006,Laguna1999}, scatter search \cite{Campos1999}, iterated local search \cite{Lourenco2002}, sorting through insertion pattern and permutation reversal \cite{Chanas1996} and evolutionary algorithms \cite{Snasel2008}.

Some evolutionary approaches to LOP presented and analyzed by Snasel et al \cite{Snasel2008} use two approaches: mutation only genetic algorithm and higher level chromosome genetic algorithm. These evolutionary methods obtain good results for problem instances from the library proposed by Mitchell and Borchers \cite{Mitchell2000}.

Two ant-based models are described for solving LOP: a hybrid model based on the Ant Colony System (ACS) \cite{Dorigo1996,Dorigo2005}, called Ant Colony System-Insert Move (ACS-IM) presented in \cite{Pintea2009} and the second model is an extended version of the Sensitive Ant Model (SAM) \cite{Chira2008} which combines stigmergic communication with heterogeneous agent behavior. This technique, called Step Back Sensitive Ant Model (SB-SAM) \cite{Chira2009a} extends SAM method by defining new (virtual state) behavior for ants having a certain sensitivity level. Lasius niger ants \cite{Baldacci2008} include u-turns in the process of selection which have a high impact on the quality of the detected paths. Inspired by Lasius niger ants behaviour, the virtual state transition rule avoids a selection of the next step from the available nodes making the agent to take a ’step back’ to the previous node and explore other regions of the search space.

The SB-SAM virtual state based decision making process generates a benefic exploratory behavior of the system which is balanced by the exploitation orientation of the highly sensitive agents in the system. 

The paper reports new results for the artificial LOP large instances data generated by Mitchell and Bochers \cite{Mitchell2000}. Both ant based techniques produce high-quality results for of LOP competing with two genetic algorithms: Mutation Only Genetic Algorithm (MOGA) and Higher Level Chromosome Genetic
Algorithm (HLCGA) \cite{Snasel2008}. The results reported by SB-SAM are better than those of ACS-IM and the already mentioned genetic algorithms. This is indicating the potential of inducing heterogeneity in ant-based models for better addressing difficult problems where complex behavior patterns are needed.

Numerical experiments on several LOP instances available in the well-known real-world data library LOLIB \cite{lolib} are reported in \cite{Chira2009b}.
This paper is organized as follows: the linear ordering optimization problem is described; ant colony models are described; numerical experiments and comparisons are discussed and directions for future research are presented. 

\section{The Linear Ordering Problem}
One of the well-known NP-hard problem is the Linear Ordering Problem (LOP). LOP is equivalent in economics with the triangulation problem for input-output tables in economics \cite{Chanas1996,Garcia2006,Reinelt1985}. Single-server scheduling, ranking by aggregation of individual preferences in group decision making, archaeological seriation are other applications \cite{Chanas1996,Laguna1999}.\\

In order to introduce LOP, let $E = (eij)_{n x n}$ be a matrix of weights. The value of $e_{ij}$ refers to the cost of having object $i$ before object $j$ in the linear ordering. 

LOP aims to find a permutation of the rows and columns  
$\pi= (\pi_{1}, \pi_{2}, \cdots , \pi_{n})$  such that the total weight (sum of elements above the diagonal) is maximized \cite{Laguna1999,Reinelt1985}:

\begin{equation}
 C_{E}(\pi)=\sum_{i=1}^{n-1}\sum_{j=i+1}^{n}e_{\pi_{i}\pi_{j}}
\end{equation}

An acyclic tournament corresponds to a permutation of the graph vertices. Let $G = (V,E,w)$ be a complete graph. V is the set of nodes and E contains the vertices of the graph and w refers to the weight (or cost) associated with an edge (the weight from $i$ to $j$ can be different that the weight of $j$ to $i$). 
The function to be maximized in LOP is the following:
\begin{equation}
 C_{G}(\pi)=\sum_{i\leq j, i\neq j}w(\pi_{i},\pi_{j})
\end{equation}

\noindent where $\pi$ is a permutation of $V $, $i, j\in V$ and $\leq$ is a total order relation on $V $.

\section{Ant-based model for solving LOP}
The ant-based models involved in the present paper to solving Linear Ordering Problem are described shortly in the following.

\begin{enumerate}[A]

\item \textit{Ant Colony Systems}\medskip

Ant Colony System (ACS) metaheuristic is introduced in \cite{Dorigo1996,Dorigo2005}. The ACS model replicates the behavior of social insects to the search space represented commonly by a graph. Each edge has an associated weight as well as a pheromone value corresponding to a desirability measure. A problem solution is a complete tour generates a by an ant. The next node in the graph is choose probabilistically based on the weight and
the amount of pheromone on the connecting edge \cite{Dorigo1996}. Stronger pheromone trails are preferred by ants and the most promising tours build up higher amounts of pheromone in time.\medskip

\item \textit{Ant Colony System-Insert Move}\medskip

The ACS model for solving LOP is called Ant Colony System-Insert Move (ACS-IM) \cite{Pintea2009}. ACS-IM starts with a greedy search. ACS-based rules are applied and a local search mechanism based on insert moves is engaged. The local search mechanism is based on the neighborhood search proposed for LOP \cite{Garcia2006}. Insert moves (IM) are used to create a neighborhood of permutations for one solution.\medskip

\item \textit{Sensitive Ant Model}\medskip

The Sensitive Ant Model (SAM) extends ACS by inducing heterogeneity in the ant population \cite{Chira2008}. 

Each ant is endowed with a pheromone sensitivity level (PSL - a real number in the unit interval).
Highly-sensitive ants are influenced by stigmergic information in the decision making process and thus likely to select strong pheromone trails. The ants with low sensitivity tend to ignore the amount of pheromone and bias the search towards unexplored regions. The new SAM probability of selecting the next node is the same with the ACS one when pheromone sensitivity level (PSL) value is maximum being reduced proportionally with the sensitivity level in the rest of the cases. A virtual state is introduced in SAM corresponding to the ’lost’ probability of (1 − PSL). The associated virtual state decision rule specifies the action to be taken when the virtual state is selected using the renormalized transition mechanism \cite{Chira2008}.\medskip

\item \textit{Step-Back Sensitive Ant Model}\medskip

The Step-Back Sensitive Ant Model (SB-SAM) \cite{Chira2009a} is based on the Sensitive Ant Model (SAM) \cite{Chira2008}. The aim of SB-SAM is to better exploit the potential of the virtual state translating to a more reliable search space exploration. This is achieved by specifying a new action associated with the virtual decision rule inspired by the observed behavior of the ant Lasius niger \cite{Baldacci2008}. Biologists’s study emphasize that some Lasius niger suddenly move in the opposite direction making a u-turn and exploiting the geometry of the trail network bifurcation \cite{Baldacci2008}. These u-turns actually contribute more to the selection of a path compared to bi-directional trail laying. SAM reinforce the inherent heterogeneity of the model facilitating a significant increase of diversity in the search. The virtual state transition rule specified by SB-SAM makes the ant to ’take a step back’ by selecting the previous node \cite{Chira2009a}. The pheromone trail is locally updated by decreasing the pheromone intensity on the edge connecting the current node with the previous one. The local virtual state update rule complements is given by the following equation:

\begin{equation}
 \tau_{ji}(t+1)=(1-\rho)\tau_{ij}-\rho\tau_0
\end{equation}
where $\tau_{ji}$ represents the amount of pheromone on edge $(j, i)$ while $\rho$ and $\tau_0$ are the same parameters used in the ACS local updating rule \cite{Dorigo1996}. The search continues from the new current node until a complete tour is built. The probability of an ant to take the virtual state decision is inverse proportionally with the ant’s sensitivity level: lower PSL means higher chance to take a step back whereas high PSL values favor the application of the ACS-inherited state transition rule.

The step-back virtual decision rule generates completely new overall behavior for the system of interacting ants triggering search diversity. A good distribution of PSL values in the population is important to create a balance in the search space between the exploitation and exploration \cite{Chira2009a}.
\end{enumerate}\medskip

\begin{enumerate}[1)]

\item \textit{Representation and initialization.}\medskip

Linear Ordering:  Problem solution is a list of vertices, of a complete directed graph with n nodes. The list is constructed in a step by step manner by each ant based on the transition rules specified by the ant model. The algorithm is based on ACS-IM \cite{Pintea2009}. It is initialized with a greedy solution obtained based on a 2-exchange Neighborhood search, permuting two positions in the ordering \cite{Resende2008}. 

A function $w$ assigns real values (the weights) to edges defining the static matrix of weights $W = (w_{ij})$, $1 \leq i, j\leq n$. The pheromone matrix $\tau$ has the same dimensions as $W$. The pheromone matrix $\tau$ is initialized with $\tau_{ij} = \tau_0$, $1 \leq i, j\leq n$ where $\tau_0$ is a small positive constant.
The SB-SAM algorithm uses m ants. A PSL value is randomly generated for each ant. The ants are initially randomly placed in the nodes of the graph. 

\begin{figure}[ht]
\begin{center}
 \includegraphics[scale=0.7]{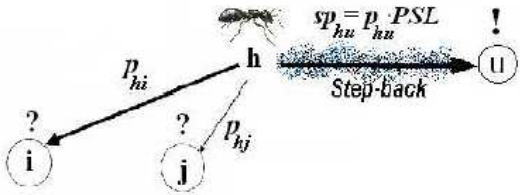}
\end{center}
\caption{Step-back technique representation}
\end{figure} 

\item \textit{Solution construction.}\medskip

The SAM renormalized transition probabilities \cite{Chira2008} guide the selection of the next node in the search process. The probability of choosing the next node $u$ from the current node $i$ for an agent $k$ is given by the following equation:
\begin{equation}
 sp_{iu}(k) = PSL(k) * p_{iu}(k)
\end{equation}
where $PSL(k)$ represents the sensitivity level of agent $k$ and $p_{iu}(k)$ is the $ACS$ state transition probability \cite{Dorigo1996}. 
The $ACS$ local update pheromone rule \cite{Dorigo1996} is applied to modify the amount of pheromone on the latter edge traversed. The probability of an ant $k$ to make a virtual state transition from a current node $i$ is equal to $(1 - \sum p_{iu}(k))$. In this case, ant $k$ goes back to the previous node and the local virtual state update rule (3) is applied. Insert moves (IM) \cite{Laguna1999} are used to improve the final solution detected by an ant. A global update pheromone rule \cite{Dorigo1996} is applied to the edges belonging to the best tour:
\begin{equation}
 \tau_{ij}(t+1)=(1-\rho)*\tau_{ij}+\frac{\rho}{C_{bs}}
\end{equation}
\noindent where $C_{bs}$ is the weight of the best-so-far solution and $\rho$ is a parameter.
\end{enumerate}

\section{New results for Mitchell and Bochers Instances (MBLIB)}
This is the main section of the paper, reporting new results for the artificial LOP data collection generated by Mitchell and Bochers \cite{Mitchell2000}. Besides comparing the performance of the two ant-based models ACS-IM and SB-SAM, numerical experiments focus on evaluating the proposed ant LOP models against Mutation Only Genetic Algorithm (MOGA) and Higher Level Chromosome Genetic Algorithm (HLCGA) \cite{Snasel2008}.

The following parameters are used for ACS-IM and SB-SAM algorithms: $\alpha= 1$, $\beta = 2$, $\tau_0 = 0.1$,$\rho = 0.1$, $q_0 = 0.5$ and $m = 10$. In the SB-SAM approach, the PSL value for each agent is randomly generated using an uniform distribution. 

Table ~\ref{table:1} shows the results for the instances available in MBLIB \cite{Mitchell2000}. The deviation of the obtained solution from the optimum solution is computed based on the average over five runs of the algorithm with 200 iterations.

\begin{table}
\begin{center}
\begin{tabular}{l l l l l l l}
\hline
Instance & Optimal  &  ACS-IM &  SB-SAM\\
\hline\\
r100a2 &197652&\bf{0.0023} &0.0027       \\
r100b2 &197423&0.0046      &\bf{0.0031}  \\
r100c2 &193952&\bf{0.0032} &\bf{0.0032 } \\
r100d2 &196397&0.0047      &\bf{0.0040}  \\ 
r100e2 &200178&\bf{0.0024 }&0.0040       \\ 
r150a0 &550666&\bf{0.0016} &0.0026       \\ 
r150a1 &504308&\bf{0.0017} &0.0023       \\ 
r150b0 &554338&0.0022	     &\bf{0.0018}  \\ 
r150b1 &500841&0.0033	     &\bf{0.0016}  \\ 
r150c0 &551451&\bf{0.0012} &0.0015       \\ 
r150c1 &500757&\bf{0.0025} &0.0025       \\ 
r150d0 &552772&\bf{0.0014} &0.0024       \\ 
r150d1 &501372&0.0039      &\bf{0.0032 } \\ 
r150e0 &554400&0.0035	     &\bf{0.0029 } \\ 
r150e1 &501422&\bf{0.0023} &0.0026       \\ 
r200a0 &989422&0.0012	     &\bf{0.0011}  \\ 
r200a1 &889222&\bf{0.0011} &0.0015       \\ 
r200b0 &984081&0.0013      &\bf{0.0012}  \\ 
r200b1 &893867&0.0013	     &\bf{0.0001 } \\ 
r200c0 &990568&\bf{0.0012} &\bf{0.0012}  \\ 
r200c1 &882945&0.0015	     &\bf{0.0001}  \\ 
r200d0 &989123&\bf{0.0013} &0.0015       \\ 
r200d1 &888563&0.0015      &\bf{0.0014}  \\ 
r200e0 &980354&0.0015	     &\bf{0.0007}  \\ 
r200e1 &883948&0.0018      &\bf{0.0010}  \\ 
r250a0 &1545431&0.0022	   &\bf{0.0021}  \\ 
r250b0 &1538410&0.0018	   &\bf{0.0017}  \\ 
r250c0 &1534036&0.0024     &\bf{0.0021}  \\ 
r250d0 &1540117&\bf{0.0017}&0.0031       \\ 
r250e0 &1531709&\bf{0.0022}&0.0043       \\ 
\hline
\end{tabular}
\caption{Numerical results of ACS-IM and SB-SAM for the $MBLIB$ instances: Deviation errors average over five runs and 200 iterations.}
\label{table:1}
\end{center}
\end{table}

\begin{figure}[htbp]
\begin{center}
 \includegraphics[scale=0.60]{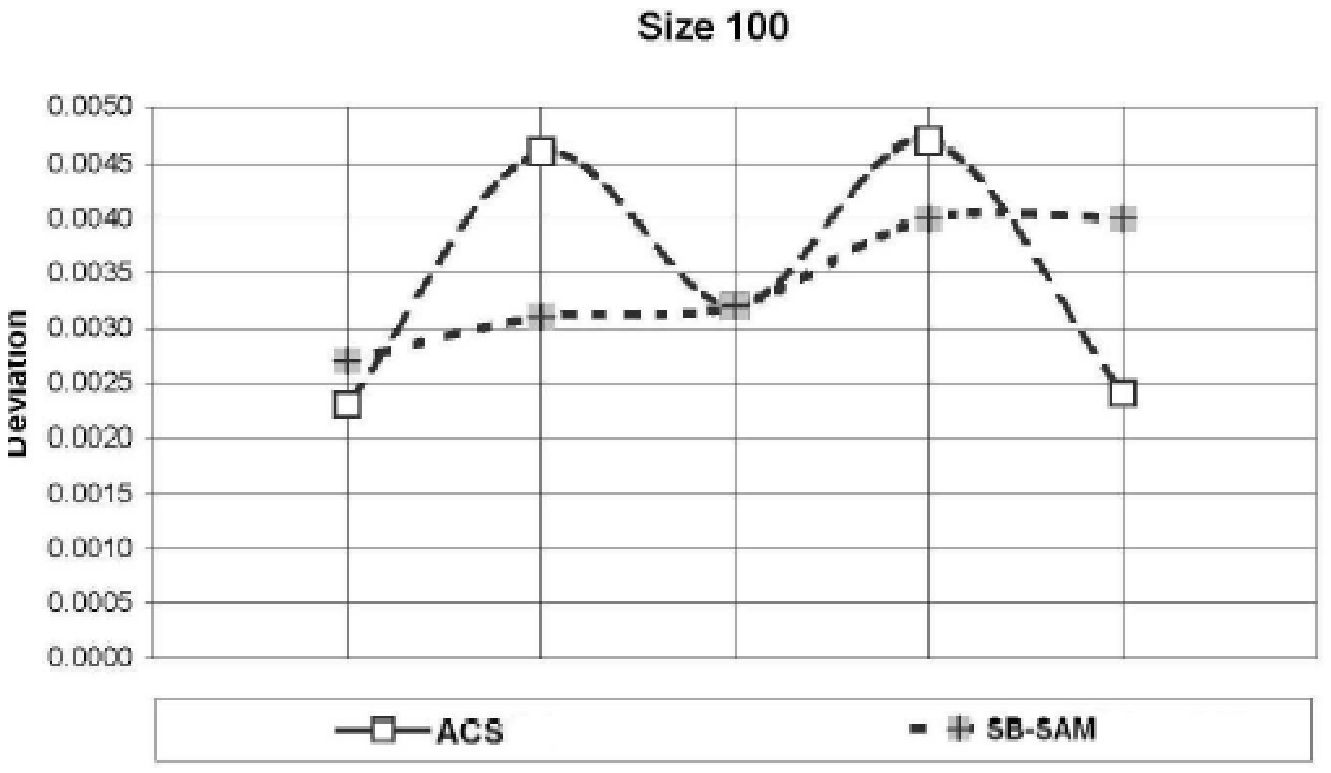}
 \includegraphics[scale=0.32]{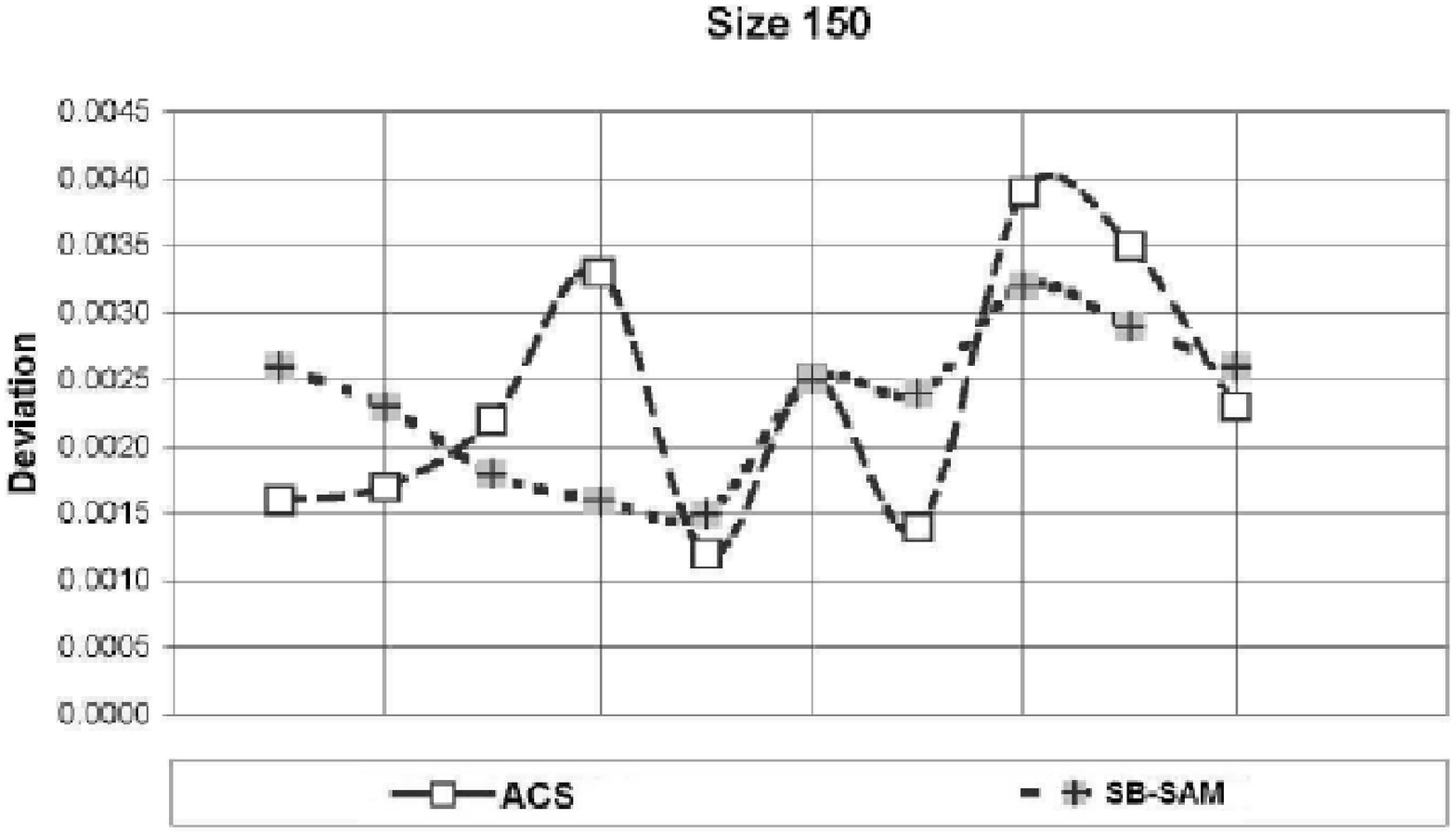}
\end{center}
\caption{Comparative ant models deviation errors: $MBLIB$ instances, size 100 and 150.}
\end{figure} 

\begin{figure}[htbp]
\begin{center}
\includegraphics[scale=0.60]{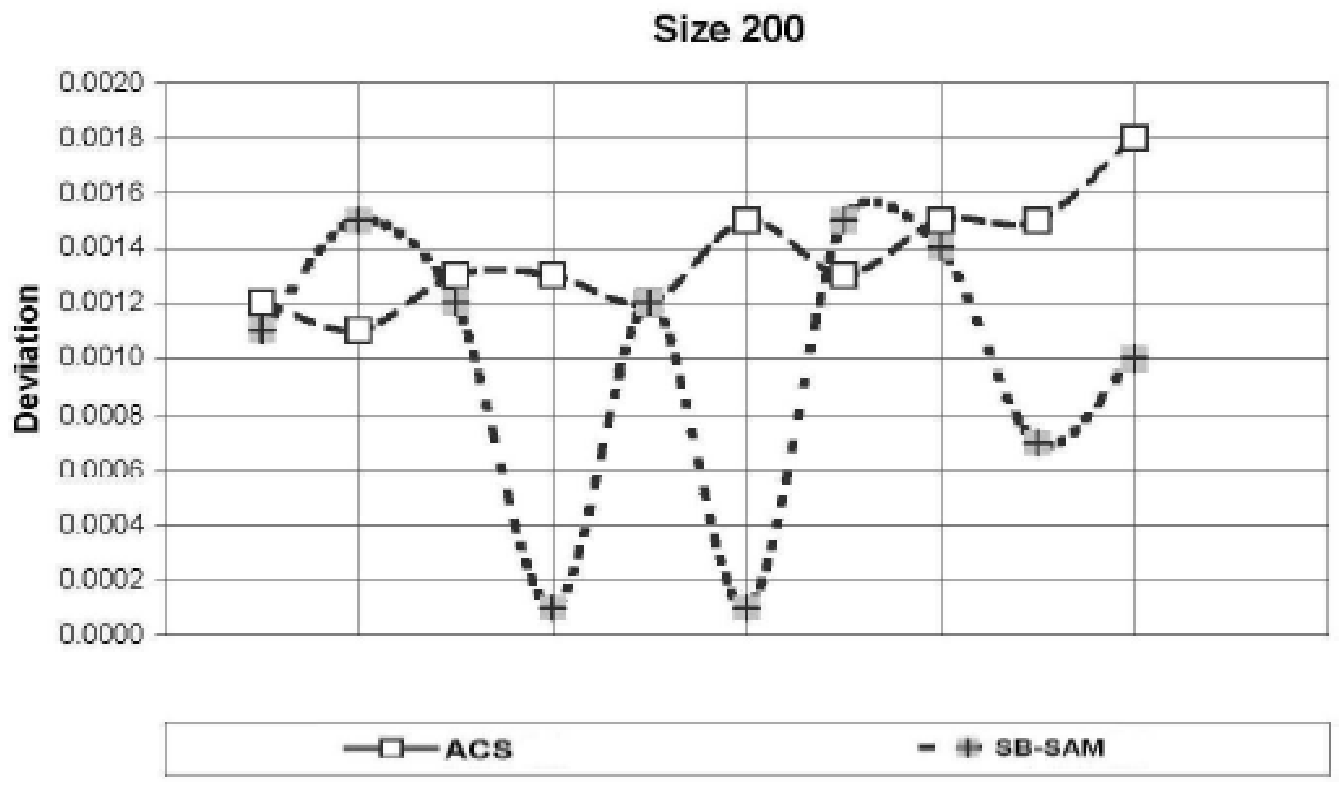}
\includegraphics[scale=0.60]{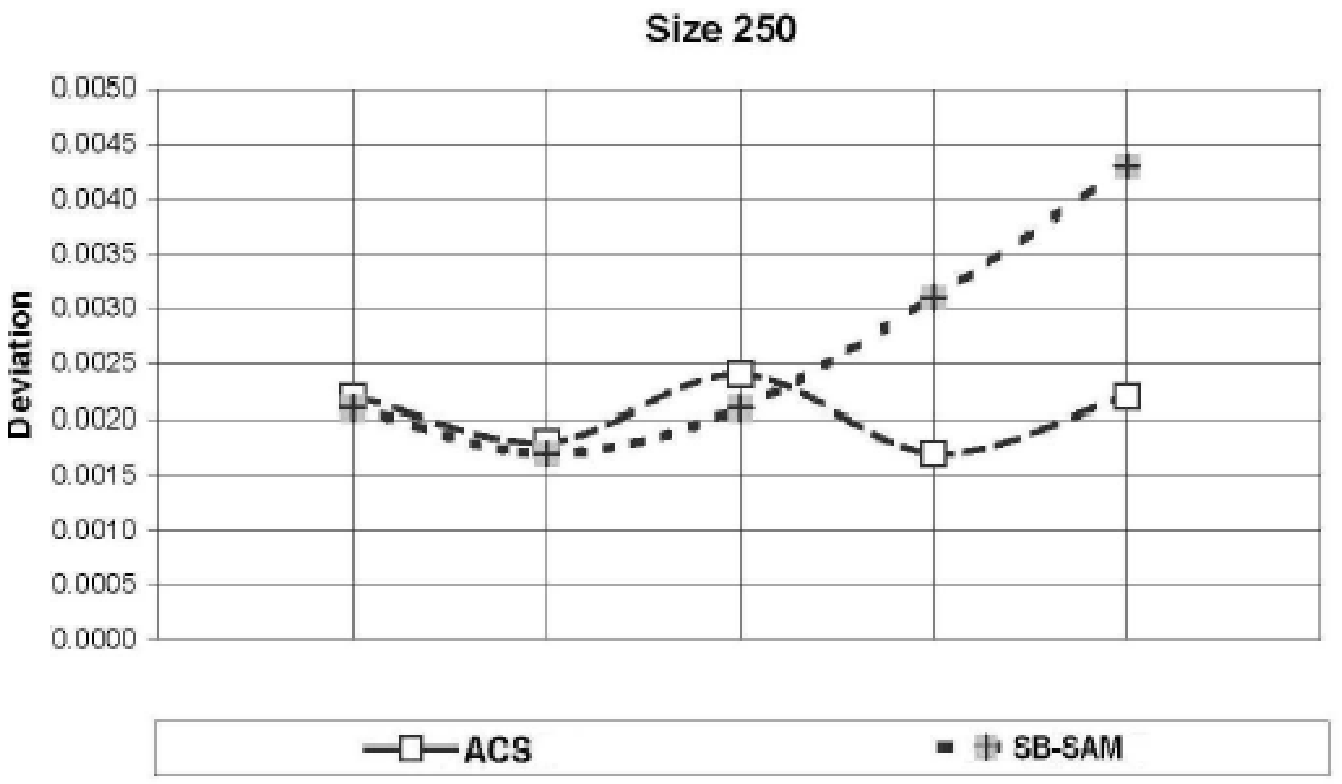}
\end{center}
\caption{Comparative ant models deviation errors: $MBLIB$ instances, size 200 and 250.}
\end{figure} 

The analysis of the results for each problem size (100-250) confirms the results obtained for MBLIB (see Figures 2-3).
SB-SAM algorithm shows an intensification of search in more complex spaces. The effects of the u-turns specified by the virtual state transition rule in SB-SAM (resulting in search diversification) are more visible for higher dimension problem instances.

\begin{table}
\begin{center}
\begin{tabular}{l l l l l }
\hline
Size&  ACS-IM& SB-SAM  & MOGA& HLCGA\\
\hline\\
100 & 0.344\%	& 0.340\%& - &-\\
150 &	0.236\% & 0.234\%& - &-\\
200 &	0.137\% & 0.098\%& - &-\\
250 &	0.206\% & 0.266\%& - &-\\
Average& 0.231\%& 0.235\%& 0.618\%&  1.606\%\\
\hline
\end{tabular}
\caption{Deviation error (in \%) of ant-based models compared to evolutionary approaches \cite{Snasel2008} for $MBLIB$ instances.}
\label{table:2}
\end{center}
\end{table}

In the following the comparison of ant-based models with Mutation Only Genetic Algorithm (MOGA) and Higher Level Chromosome Genetic Algorithm (HLCGA) \cite{Snasel2008}. Average overall results for MBLIB instances are available for these algorithms in \cite{Snasel2008}. Table~\ref{table:2}  shows the average deviation obtained by all involved algorithms for MBLIB instances. In \cite{Garcia2006} are proposed also some hybrid models combining variable neighborhood search and different strategies for local search. Local search strategies are limited by the dimension of problem instances.

Further work will focuses on other hybrid ant-based techniques in order to obtain better solution for large LOP instances. 

\section{Conclusions}
The paper shows new results for Linear Ordering Problem (LOP) using bio-inspired models. The ant-based models for solving LOP are described: the first one is a hybrid Ant Colony System \cite{Dorigo2005}, called Ant Colony System-Insert Move (ACS-IM) \cite{Pintea2009} and the second one uses a new Step-Back Sensitive Ant model \cite{Chira2008,Chira2009a}. The ant models compete with other techniques for solving LOP in terms of solution quality.

Numerical experiments indicate a better performance of SBSAM in solving large Mitchell instances \cite{Mitchell2000}, compared to the ACS-IM technique and two genetic algorithms: Mutation Only Genetic Algorithm (MOGA) and Higher Level Chromosome Genetic Algorithm (HLCGA) \cite{Snasel2008}. Other hybridization techniques for solving NP-hard problems, including Linear Ordering Problem will be developed.

\section*{Acknowledgement}
This research is supported by the CNCSIS Grant ID 508 New Computational Paradigms for Dynamic Complex Problems funded by the Ministry of Education, Research and Innovation, Romania.

\end{document}